\documentclass[12pt]{amsart}
\usepackage{amsbsy,amssymb,amsmath,amsthm,amscd,amsfonts,latexsym,amstext,delarray,
amsmath,graphicx} 
\usepackage[margin=.8in]{geometry}
\usepackage{color}

\numberwithin{equation}{section}

\def\Z{{\mathbb Z}}

\def\P{{\mathbb P}}

\def\cL{{\mathcal L}}

\def\fA{{\mathfrak A}}

\title{Spin Glass Models of Syntax and Language Evolution}
\author{Karthik Siva, Jim Tao, Matilde Marcolli}
\address{Division of Physics, Mathematics, and Astronomy \\ California Institute of Technology \\
1200 E. California Blvd, Pasadena, CA 91125, USA}
\email{ksiva@caltech.edu}
\email{jtao@caltech.edu}
\email{matilde@caltech.edu}
\date{}

\begin{document}
\maketitle

\begin{abstract}
Using the SSWL database of syntactic parameters of world languages, and the MIT Media Lab
data on language interactions, we construct a spin glass model of language evolution. We treat  
binary syntactic parameters as spin states, with languages as vertices of a graph, and assigned
interaction energies along the edges. We study a rough model of syntax evolution, under the
assumption that a strong interaction energy tends to cause parameters to align, as in the case
of ferromagnetic materials. We also study how the spin glass model needs to be modified
to account for entailment relations between syntactic parameters. This modification leads
naturally to a generalization of Potts models with external magnetic field, which consists of
a coupling at the vertices of an Ising model and a Potts model with $q=3$, that have the same 
edge interactions. We describe the results of simulations of the dynamics of these models,
in different temperature and energy regimes. We discuss the linguistic interpretation
of the parameters of the physical model.
\end{abstract}

\section{Introduction}

The evolution of languages through the interaction of their speakers is a topic of interest to computational
linguists and, like many interacting many-body problems, is difficult to study analytically. In this paper we
follow an approach that views languages at the level of syntax, with syntactic structures encoded as a string of binary syntactic parameters, a point of view originating in the Principles and Parameters model
of Generative Linguistics, \cite{Chomsky}, \cite{ChoLa} (see also \cite{Baker} for a more 
expository account). It is known that syntactic parameters can change
in the course of language evolution. Cases of parameter flipping have been identified in the historical development
of some Indo-European languages, see for example \cite{Taylor}. For recent results on language evolution
from the point of view of syntactic parameters, see \cite{Galv}. 

\smallskip

We construct a model for language evolution
 inspired by the physics of {\em spin glasses}. These are systems of interacting spin variables, with spins
 located at the vertices of a graph and with interaction energies along the edges that tend to favor alignment 
 (ferromagnetic) or anti-alignment (anti-ferromagnetic) of the spin variables at the endpoints of each edge. 
 The dynamics of the system also depends on thermodynamic temperature parameter, which is a measure 
 of disorder in the system, so that the spin variables tend to be frozen onto the ground state at low temperature, 
 while higher energy states become accessible to the dynamics at higher temperature. 
 
 \smallskip
 
We interpret each syntactic parameter as a spin variable, on a graph representing world languages and their interaction. 
We obtain the binary values of the syntactic parameters from the Syntactic Structures of the World's Languages
(SSWL) database\footnote{{\tt http://sswl.railsplayground.net/}}, which documents these values for 111 
syntactic parameters and over 200 natural languages. To model the interaction strengths between
languages, we use data from the MIT Media Lab\footnote{{\tt http://language.media.mit.edu/visualizations/wikipedia}}, by defining the strength of the influence of language A on language B as the likelihood 
that two languages are to be co-spoken. In particular, in their database, two languages
are connected when users that edit an article in one Wikipedia language edition are significantly more
likely to also edit an article in another language edition. The frequency of such occurrences provides
an estimate of the strength of the interaction.
 
 \smallskip

The idea of modeling syntactic parameters as spin variables in a statistical mechanical setting is
not entirely new to Computational Linguistics. A model based on this idea was proposed in \cite{Niyogi}.
The main difference with respect to the approach we follow here is that, in the model of \cite{Niyogi}, 
the vertices of the graph are individual speakers in a fixed population, rather than languages 
(populations of speakers)  as in our model. A statistical physics model of language change based 
on syntactic parameters was also constructed in \cite{CCGG}.

\medskip
\subsection{General assumptions of the model}

We make a series of simplifying assumptions, to the purpose of obtaining a computationally feasible model.
We will examine the plausibility of these assumptions and their interpretation from a Linguistics point of view. 

\smallskip

First, we assume that the languages we
simulate are sufficiently distinct and never converge and do not concern ourselves with, for example, whether
a dialect of language A is truly distinct from language A or whether two closely related languages A and
B will at some point just become the ``same" language. Instead, we assume there exists a definition of a
language for which the notion of distinct languages is precise and for which the languages we have identified
are always distinct. 

\smallskip

The second simplification we make is that for a given syntactic parameter, such as the
``Subject-Verb" syntax, a language either has it or does not have it. One could account for finer syntactical
structures by considering syntaxes of arbitrary length, but this would still admit a binary classification
over the languages. 

\smallskip

A third assumption is that because language
interaction occurs microscopically by human interaction, and a foreign language is easier to acquire if its
syntax is familiar, interacting languages will generally prefer to align their syntactic parameters. From these
assumptions, we construct a Hamiltonian and evolve the system from its current syntactic parameter state
toward equilibrium.

\medskip
\subsection{Discussion of assumptions}

Considering languages as ``discrete" objects (as opposed to a continuum of dialects) is a rather
common linguistic assumption. Alternative models, such as wave models of transmission of linguistic
changes are also possible (see for example \cite{Nerb}), but we will not consider them in this
paper. It would be interesting to see whether statistical physics methods could be relevant to
wave models of languages, but that is outside the purpose of our present investigation. 

\smallskip

The second assumption listed above is clearly more problematic: it is a drastic simplification, 
which ignores phenomena of entailment between syntactic parameters. Indeed, it is well known
that there are relations between different syntactic parameters, while our assumption leads us
to treat them as independent spin variables. For example, there are pairs of parameters
$(p,p')$ with the property that if $p=+1$ then $p'$ is undefined, while if $p=-1$,
then $p'$ can take either value $\pm 1$: see \cite{Longo1}, \cite{Longo2} for some explicit
examples of this entailment behavior. Thus, in a more refined version of the model,
the second assumption above should be modified in two ways: (1) an additional possible value
$0$ of the parameters should be introduced, which accounts for the case where a parameter
is undefined; (2) relations between parameters should be
introduced modeling the entailment property described above. The first modification simply
corresponds, in spin glass models, to considering Potts models with number of possible
spin states $q=3$, instead of Ising models with $q=2$ (see \cite{Sokal}
for a survey of the mathematical properties and differences between these models), 
while the second modification requires relations (at the vertices of the graph) between
different spin variables (syntactic parameters) and can be modeled on versions of
Potts models with external magnetic fields, \cite{ElMoMoff}.  We will first consider
the simpler version of the model, with the second assumption as listed above, and
then discuss the effect of introducing modifications as indicated here. The case with
entailment of parameters is discussed in \S \ref{entailSec}. A simple modification of
the entailment model of \S \ref{entailSec} also account for variants of the entailment
relation (see \cite{Longo1}, \cite{Longo2}) where $p=-1$ implies $p'$ undefined, 
and $p=+1$ implies $p' =\pm 1$, or where the parameter $p_2$ in entailed by
more than one parameter. The latter case requires the coupling at vertices of more than
two spin glass models. We will focus on only one case for simplicity.

\smallskip

Regarding the third assumption, since our model of interaction energies for the
spin glass system is based on measures of bilingualism, as obtained by Media Lab,
the evolution of syntactic parameters that we describe would be best understood,
linguistically, within the context of the theory of bilingual Code-switching, \cite{MacSwan}.
For example, as observed in the study of \cite{Wool} of English--Spanish bilingual population,
the language spoken by this population ``acquires" the Pro-Drop parameter (which is $+1$
in Spanish but $-1$ in English) in the sense that Spanish verbs inserted in English sentences
retain their Pro-Drop form from Spanish.  In the evolution of
syntactic parameters we obtain from the spin glass simulation, we see previously
inactive $-1$ parameters switch to the activated form $+1$ as an effect of ``ferromagnetic"
alignment with the corresponding parameters of nearest neighbor vertices. In a realistic model, 
this should be interpreted (at least in a first phase of language evolution)  in the sense of Code-switching
theory, as imported forms from the neighboring languages that are inserted in the original
syntactic structure, while retaining their parameters activated. 

\smallskip

Finally, we should comment on our overall choice of describing languages in
terms of syntactic parameters, within the Principles and Parameters setting of
\cite{Chomsky}, \cite{ChoLa}. While this framework can been criticized, see
for example \cite{Hasp2}, it is clear that coding syntactic structures as a string
of binary variables makes it extremely suitable for modeling via statistical physics 
methods, hence it is a very natural setting for our purposes.
We proceed to a more detailed description of the model in the coming section.

\medskip 

\section{Spin Glass Model}

The general data of a spin glass model (Potts model) consist 
of a finite graph $G$; a finite set $\fA$
of possible spin states at a vertex, with $\# \fA=q$; a set $\Sigma={\rm Maps}(V(G),\fA)$
of possible states of the system, namely assignments of a spin state in $\fA$ at each
vertex in $V(G)$; a set of interaction energies associated to the edges $e\in E(G)$.
Usually the energies are assumed to be equal to zero is the spins at the endpoints 
$v\in \partial(e)$ are not aligned and equal to a value $-J_e$ if they are aligned.
The ``ferromagnetic" case corresponds to the assumption that all $J_e\geq 0$ and
the ``antiferromagnetic" case corresponds to $-\infty \leq J_e\leq 0$. One often expresses
the interaction energies in terms of edge variables $t_e = e^{\beta J_e}-1$, where 
$\beta$ is a thermodynamic parameter (an inverse temperature).  

\smallskip

In the Ising model case, where spin variables have two possible values $\pm 1$,
a typical form of the Hamiltonian of a spin glass model is, for states $S\in \Sigma$
\begin{equation}\label{HamIsingB}
H(S) =- \sum_{e\in E(G): \partial(e)=\{ v,v'\}} J_e\, S_v S_{v'} - \sum_{v\in V(G)} B_v \, S_v,
\end{equation}
where $B=(B_v)$ is an external magnetic field: the first term measures the degree of
alignment between nearby spins and the second the alignment of spins with the direction
of the overall external field. When we allow for more general Potts model cases, where the
spin variables have $q\geq 2$ possible values, Hamiltonian \eqref{HamIsingB} is rewritten as
\begin{equation}\label{HamPottsB}
H(S) =- \sum_{e\in E(G): \partial(e)=\{ v,v'\}} J_e\, \delta_{S_v, S_{v'}} - \sum_{v\in V(G)} B_v \, S_v .
\end{equation}
The partition function of the system is
\begin{equation}\label{PartitionPottsB}
Z_G(\beta)=\sum_{S \in \Sigma} \exp(-\beta H(S)),
\end{equation}
and the associated Gibbs probability measure is
\begin{equation}\label{Gibbs}
\P_{G,\beta}(S) = \frac{e^{-\beta H(S)}}{Z_G(\beta)}.
\end{equation}
For more general cases, and a more detailed mathematical 
discussion see \cite{ElMoMoff}, \cite{Sokal}. In particular, one can fit into this
model the case where there are two different interaction energies $J_{e}\neq J_{\bar e}$
associated to the two opposite orientations of $e$, by doubling all edges in the graph,
so as to consider both orientations, with respective energies $J_e$ and $J_{\bar e}$.
This case will be relevant to our application.

\medskip
\subsection{Topology and Energetics of the System}

With these simplifications and resources, the stage is set to map the problem onto a classical statistical mechanics problem. We identify each language with the vertex of a digraph, and each interaction of the form {\em language $A$ influences language $B$} with a directed edge from vertex A to vertex B. Such an edge has a weight or interaction strength $J_{AB}$ given by the aforementioned metric used by MIT Media Lab. Note that in general, $J_{AB}\neq J_{BA}$. 

\smallskip

We represent the presence or absence of a particular syntactic parameter by associating 
with each vertex a spin-$1/2$ variable, where a spin of $+1/2$ indicates the former and $-1/2$ represents the 
latter. For simplicity, in the following we just use either $+1$ or $-1$ for the two possible spin states, as described above, 
referring to them as ``spin-up'' and ``spin-down'', respectively, 
absorbing a factor of $1/2$ into the term $J_{\ell\ell'}$. 
The physical analog of this setup is the association of a particle with spin attributes to each vertex and 
inter-particle interactions with the edges, the usual setting of Ising models in statistical physics. 

\smallskip

The full specification of the configuration of the system in the 
single-particle (single-language) basis is then given by
\begin{equation}\label{singleL}
|\vec{S}\rangle = \bigotimes_{p\in\mathcal{P}}|\vec{S_p}\rangle =  \bigotimes_{p\in\mathcal{P}} \left(\bigotimes_{\ell \in \mathcal{L}} |S_{\ell,p}\rangle\right)
\end{equation}
where $S_{\ell,p}\in\{-1,+1\}$, $\mathcal{P}$ is the set of all parameters, and $\mathcal{L}$ is the set of all languages. 

\smallskip

With this picture in mind, we turn to the evolution.  In this tensor product basis, we
can simulate the evolution of each of the 111 syntactic parameters, in the {\em independent parameters
approximation}, by endowing the system with an interacting Hamiltonian for each syntactic
parameter:
\begin{equation}\label{Hindep}
 H_p = -\sum_{\ell,\ell ' \in\mathcal{L}} J_{\ell\ell '} S_{\ell,p} S_{\ell ',p} .
\end{equation}
This Hamiltonian is minimized by the alignment of adjacent spins, as $J_{\ell\ell '} \geq 0\quad\forall \ell,\ell '$.  
Note that this is not the most general form of spin glass models, where the Hamiltonian 
typically contains additional disorder terms. We are also not considering here the 
possible presence of terms playing the role of an external magnetic field B, as in 
\eqref{PartitionPottsB}.

\medskip
\subsection{Dynamics and Equilibrium Physics}

We now need to impart dynamics into the model by introducing fluctuations or thermal noise.
Continuing in the physical analogy, this system at thermal equilibrium with a ``heat bath" at 
temperature $T$ will have a distribution of configurations specified by the probability density function
\begin{equation}\label{pdf}
\P(|\vec{S_p}\rangle)=\frac{1}{Z(\beta)}\exp(-\beta H_p(|\vec{S_p}\rangle))
\end{equation}
where $Z$ is the normalization or partition function
\begin{equation}\label{pf}
Z(\beta)=\sum_{\{|\vec{S_p}\rangle\}}\exp(-\beta H_p(|\vec{S_p}\rangle))
\end{equation}
and $\beta=T^{-1}$ an inverse temperature parameter (or $(k_BT)^{-1}$ in Physics, 
with $k_B$ the Boltzmann constant). 

\smallskip

Once again, let us translate this to Linguistics. The temperature
$T$ defines the extent to which languages are ``noisy'' due to external factors (the heat bath); if they are
individually very noisy, then we can expect that these fluctuations will, on average at equilibrium, result in
no overall alignment of a syntactic parameter (magnetization of zero). In this limit, the languages fluctuate
independently. On the other hand, at $T = 0$, all of the noise is frozen out, so the system will be found only
in the configuration with the largest probability as per equation~\eqref{pdf}. That is the ground state, 
or the state with minimum $H_p$, in which all languages either possess or lack a given parameter 
(magnetization of $\pm1$).

\medskip
\subsection{Temperature parameter}\label{tempSec}

We comment more in detail here on the interpretation, in Linguistics, of the temperature
parameter of the spin glass model. As we mentioned above, we want to interpret that
as a heat bath that introduces a certain level of noisiness in the behavior of the
parameters. In Linguistics, a new probabilistic approach to syntactic parameters was recently
developed in \cite{Liu}.
It makes a strong case, based on an analysis of treebanks of different sizes for a sample
of 20 languages, for viewing the syntactic parameters of a given language not as frozen
on the up or down position, $\pm 1$, but as a binary probability distribution $\{ P, 1-P \}$,
which expresses the tendency of the language of have a given syntactic parameter expressed
or not. For example, the case of the Head-Initial structure is analyzed in \cite{Liu} and it
is shown that languages typically tend to express elements of both $\pm 1$ possibilities
for this parameters, with a certain probability. Similar results are obtained, in the same paper, 
for Subject-Verb, Object-Verb, and Adjective-Noun parameters. 

\smallskip

In the light of this approach, we can regard each language $\ell$, and each
parameter $p$, as endowed with the additional data of probability distributions
$\{ P_{\ell.p}, 1-P_{\ell,p} \}$, which represent how ``noisy" the setting of the
parameter $p$ is in the language $\ell$. 

\smallskip

In general, one expects that the probability distributions $\{ P_{\ell.p}, 1-P_{\ell,p} \}$
will be different for different languages $\ell$ and different parameter $p$. However,
we can assume that they all depend on an overall thermodynamic parameter $\beta$
that can be tuned to increase or decrease the amount of noise. 
In first approximation, one can assume for simplicity that all probabilities are the
same, with a form like $P_{\ell,p}(\beta)=P(\beta)=1-\frac{e^{-\beta}}{2}$, so that, at zero temperature
($\beta=\infty$) the parameter $p$ is set to $+1$ without any noisiness (or to $-1$ in the
symmetric case), while for $T\to \infty$ (at $\beta=0$) one has the maximum amount of
noise, with the uniform distribution $P=1/2$ assigning equal probability to the two 
$\pm 1$ values of the syntactic parameter. 

\smallskip

One can consider more general functions $P_{\ell,p}(\beta)$, which are continuous
and monotonically interpolating between the values $P_{\ell,p}(\infty)=1$
and $P_{\ell,p}(0)=1/2$. For example, one can slightly modify the case discussed
above by taking $P_{\ell,p}(\beta)=1-\frac{e^{-\beta \gamma_{\ell,p}}}{2}$, where the
exponents $\gamma_{\ell,p}$ should be fitted to data like those collected in \cite{Liu},
so that there is some common value $\beta_0$ at which the probability
distribution $P_{\ell,p}(\beta_0)$ match the statistics for specific syntactic parameters
in specific languages. Unfortunately, at present, we do not have a sufficiently large set
of data, of the type collected in \cite{Liu}, to fully implement this analysis,
so we only outlined here the general approach.

\medskip
\subsection{Variant of the Model: Entailment of Parameters}\label{entsec1}

In the form of the spin glass model described here above, we made the overall 
simplifying assumption of treating all syntactic parameters as mutually independent
and uncoupled. Thus, the spin glass model is simply an uncoupled collection
of many independent Ising models, with the same underlying graph topology and
interaction energies, one for each parameter $p$.

\smallskip

This ``independent parameters assumption" is the most problematic. 
phenomena of entailment of parameters are well known, see \cite{Longo1}, \cite{Longo2}
for specific examples. In such cases,  what typically happen is that one or more parameters
depend on the setting of another parameter (or more). As a typical example, we can
take the case of two parameters $(p_1,p_2)$ with the property that, if the parameter
$p_1$ is activated (set to $+1$ value) in a language $\ell$, then the second parameter
$p_2$ can take either value $+1$ or $-1$, while if $p_1$ is set to $-1$, then the parameter
$p_2$ is just undefined in that language. Several such examples, in the family of
Indo-European languages, are reported in the table of syntactic parameters given in 
Table A, Figure 1, in the Appendix of \cite{Longo1}. We will focus on the analysis
of one such example in \S \ref{entailSec}.

\smallskip

In addition to the entailment problem, a more general problem, which is less well
understood, is whether there are other dependencies between different syntactic
parameters, and what would constitute a minimal number of independent binary
variables. This is part of the general critique that is addressed at the Principles
and Parameters model, \cite{Hasp2}. For the purpose of the present paper, we
will ignore this further problem, and we will assume that, except for the entailment
phenomena described above, the syntactic parameters can be considered 
independent and uncoupled.

\medskip
\subsection{Comparison with other models}

A statistical physics model of language acquisition and language evolution was developed in \cite{CCGG}.
Their model is also based on the Principles and Parameters approach. However, the statistical physics
model they construct is significantly different from the one we describe in the present paper. They are
mostly interested in modeling language acquisition. They model by a Gibbs state the probability with which
a speaker selects sentences, so that the choice should follow euphonic considerations: how best 
a given sentence fits the prosodic patterns of the language. To this purpose, 
they model prosody as a potential, which defines the Hamiltonian of the Gibbs state.
The grammar is then estimated by a maximum likelihood argument maximizing the Gibbs 
probability measure.

\smallskip

Another linguistic application of statistical physics models, and in particular of the Ising 
spin glass model, occurs in \S 13.6.2--13.6.3 of \cite{Niyogi}. This is a model of language
evolution, where the author considers a population of linguistic agents, corresponding to
the vertices of the graph, with edges representing the interaction between individual agents.
The model also assumes that there are two distinct languages in the population. The adoption
of one or the other language by a given speaker corresponds to the two possible spin states
$\pm 1$ at that vertex in the spin glass system. The inverse temperature parameter of the Ising
model is interpreted as the probability with which speakers of each language produce ``cues"
(in a cue-based learning algorithm). The behavior of the system then models a
situation where, above a certain critical temperature, individual agents behave independently
and both languages are equally represented in the population, while below a critical temperature
one language becomes dominant. 

\smallskip

To our knowledge, a model like ours, where the syntactic parameters themselves 
are treated as spin variables in a spin glass model, and interactions are modeled
by bilingualism data, has not been previously considered. 

\begin{figure}
\begin{center}
\includegraphics[scale=1.1]{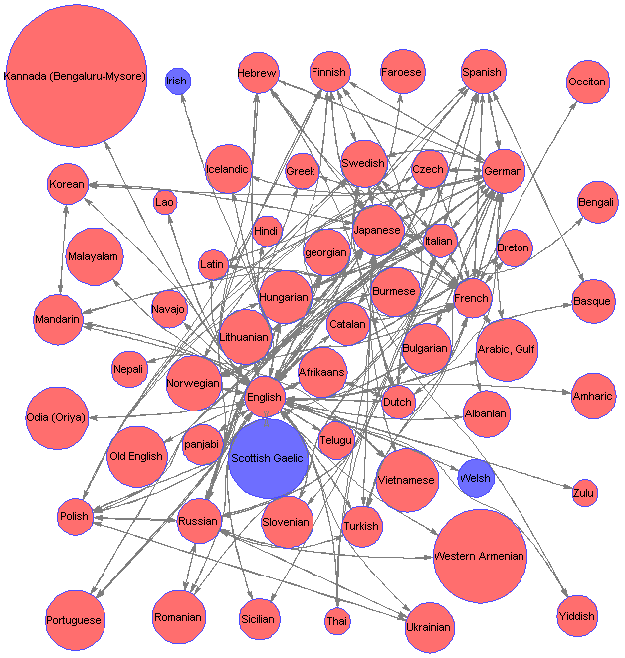}
\end{center}
\caption{Initial state of the Subject-Verb parameter for various languages in the database. 
Most languages currently possess this syntax (value $+1$ of the parameter, red colored vertices).
Graph built on MIT Media Lab data of language interaction.}
\label{init}
\end{figure}

\begin{figure}
\begin{center}
\includegraphics[scale=1.1]{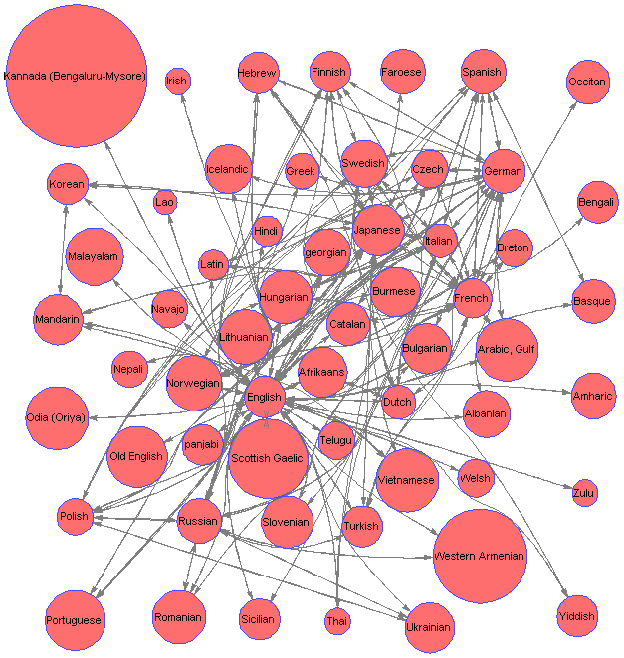}
\end{center}
\caption{In the low temperature regime $T = 0.000001$ equilibrium, all of the 
languages are expected to acquire the Subject-Verb parameter in the $+1$ position.}
\label{smallt}
\end{figure}

\begin{figure}
\begin{center}
\includegraphics[scale=1.1]{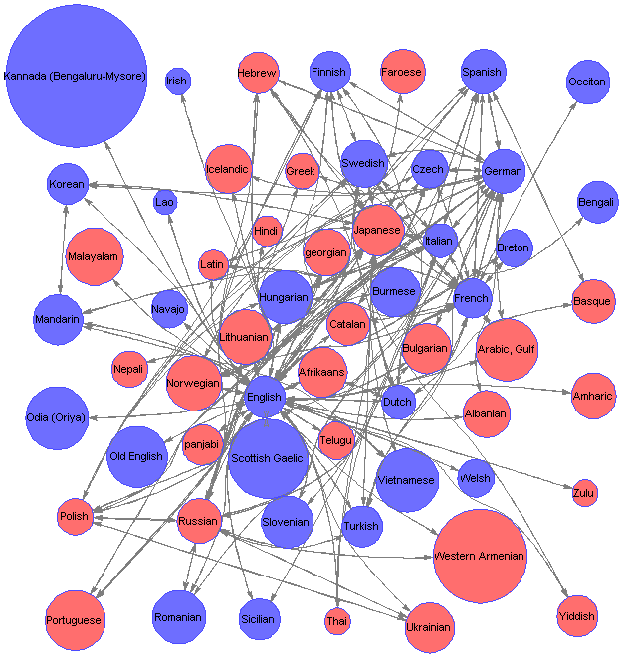}
\end{center}
\caption{In the high temperature regime $T = 20$ equilibrium for the Subject-Verb parameter, 
all vertices have local magnetization close to zero, with half of them approaching zero from 
the positive direction.}
\label{larget}
\end{figure}

\section{Computational approach}

Having established the equilibrium physics expected of a spin glass model of syntax, we
seek to explore the evolution of the system computationally. 
Naively exploring the configuration space is computationally too 
expensive, so instead we employed a Markov Chain Monte Carlo simulation to propagate stochastically an
initial configuration of syntactic parameter values (as specified by the SSWL) toward equilibrium, through the
Metropolis algorithm, see for instance \cite{Krauth}, \cite{NewBark}.

\smallskip

The algorithm (see, for instance, \S  2 and 3 of \cite{NewBark}) is based on the 
detailed balance condition
\begin{equation}\label{Pdetbal}
\P(s) \P( s\to s') = \P(s') \P(s'\to s),
\end{equation}
where $\P(s\to s')$ is the probability of transitioning from a state $s$ to a
state $s'$, represents the condition that $\P(s)$ is the stationary distribution
of a Markov process with transition probabilities $\P(s\to s')$. Uniqueness
of the stationary distribution follows from an ergodicity property of
the Markov process, usually implied by conditions of aperiodicity and 
positive recurrence. 

\smallskip

In the Metropolis--Hastings algorithm, a Markov process with the required
properties is constructed by obtaining transition probabilities $\P(s\to s')$
in terms of acceptance-rejection rates:
\begin{equation}\label{Paccrej}
\P( s\to s') = \pi_A ( s\to s' ) \cdot \pi( s\to s' ), 
\end{equation}
where $\pi( s\to s' )$ is the conditional probability of proposing a state $s'$ given the state $s$
and $\pi_A ( s\to s' )$ is the conditional probability of accepting the proposed state $s'$ given $s$.
The $\P( s\to s')$ are normalized to add up to one, by requiring that, in the remaining cases
the proposed state is $s'=s$. The usual Metropolis--Hastings choice of the acceptance
distribution is
\begin{equation}\label{MHacc}
\pi_A ( s\to s') = \min \{ 1, \frac{\P(s')\, \pi(s'\to s)}{\P(s')\, \pi(s\to s')} \} 
\end{equation}
which satisfies the balance condition \eqref{Pdetbal}
$$ \frac{\pi_A(s\to s')}{\pi_A(s'\to s)} = \frac{\P(s')\, \pi(s'\to s)}{\P(s')\, \pi(s\to s')}. $$

\smallskip

The selection probabilities $\pi(s\to s')$ in the Metropolis--Hastings algorithm
are chosen so that the ergodicity condition holds. In the case where the graph is a
lattice, this is usually achieved by considering single-spin-flip dynamics, where in each
allowed transition $s\to s'$ the final state $s'$ differs from the initial state $s$ by the flipping
of a single spin. On these possible states, the probability is taken to be uniform,
$\pi(s\to s')=\frac{1}{N}$, with $N$ the number of allowed states. 

\smallskip

In the case where $\P$ is the Gibbs measure of a spin glass model, a choice of the
acceptance probabilities $\pi_A(s \to s')$ that satisfy the detailed balance condition \eqref{Pdetbal}
is given by setting
\begin{equation}\label{piAGibbs}
\pi_A(s \to s')=\left\{
\begin{matrix}
1 & \text{ if }  H(s')-H(s) \le 0 \\
\exp(-\beta (H(s')-H(s))) & \text{ if } H(s')-H(s) >0.
\end{matrix}\right.
\end{equation}
Namely, a new state $s'$ with energy lower than or equal to that of $s$ is automatically accepted,
while one with a higher energy is accepted with probability $\exp(-\beta (H(s')-H(s)))$.
See \S 3.1 of \cite{NewBark} for more details.

\begin{figure}
\begin{center}
\includegraphics[scale=0.65]{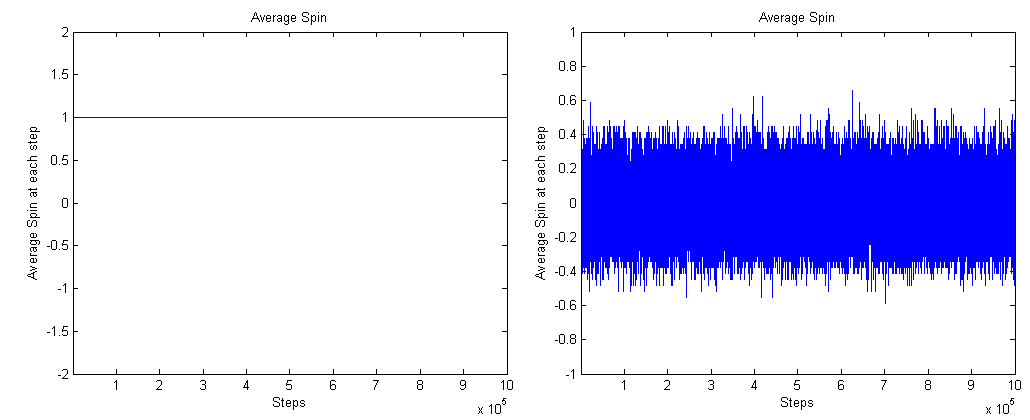}
\end{center}
\caption{Average (over vertices) value of spin, in the low temperature regime $T = 0.000001$ (left) and in the high temperature range $T = 20$ (right), as a function of the number of steps in the Monte Carlo simulation.}
\label{avgspinFig}
\end{figure}

\medskip
\subsection{Ergodicity and Metropolis--Hastings algorithm}

The ergodicity property is necessary for the working of the Metropolis--Hastings algorithm.
A Markov chain is ergodic if and only if it is irreducible and aperiodic. Irreducibility is the 
condition that the Markov process can reach any possible state of the system
starting from any given state in a finite number of steps. A state is periodic if the chain can
return to it only at multiples of some period $d>1$. Aperiodicity is the absence of periodic states
($d=1$ for all states). Under the assumption of single-spin-flip dynamics, 
ergodicity is satisfied for the Ising model on a lattice (see \S 3.1 of \cite{NewBark}). 

\smallskip

The graph we consider here is not a lattice. However, it makes sense in terms
of the linguistic interpretation of the model, to assume a similar type of single-spin-flip dynamics.
For similar arguments in favor of single-spin-flip dynamics in models of language
acquisition and evolution, see \cite{Niyogi}.  Ergodicity of the Metropolis--Hastings algorithm, 
with the single-spin-flip dynamics, still holds for our graph. The following simple argument applies
more generally.

\smallskip

Irreducibility can be checked using the notion of accessibility. Given states $s$ and $s'$,
let $\tau$ denote the minimum number of steps needed for the Markov chain to reach $s'$
starting at $s$. The state $s'$ is accessible from $s$ if there is a positive probability that
$\tau <\infty$. Suppose that the states $s$ and $s'$ differ at a certain set $A\subset V(G)$ of
sites, with $k=\# A$ and $N=\# V(G)$. Then $\P(\tau <\infty)\geq \P(\tau =k)$. Let $\cL_A$
be the set of all $k!$ orderings of the elements of $A$. For $j=1,\ldots, k!$, and $\sigma_j\in \cL_A$,
let $s_j=(s_{j,r})_{r=0\ldots,k-1}$ be the corresponding list of $k$ states in $\Sigma$, each
obtained from the previous one by flipping the spin located at the site $\sigma_j(r)\in A$. 
The probability $\P(\tau =k)$ is then the sum of the  
products of the probability of obtaining a string $\sigma_j$ in $\cL_A$ by choosing
uniformly randomly a sequence of $k$ elements in $V(G)$, times the 
product of the probabilities of the 
flipping of the individual spins at the sites $\sigma_j(r)$ for $r=1,\ldots,k$,
$$ \P(\tau =k) = \frac{1}{N!} \sum_{j=1}^{k!} \prod_{r=0}^{k-1} \pi_A(s_{j,r}\to s_{j,r+1}). $$
The precise values of these probabilities depends on the topology of the graph $G$, through
the set of edges, that determine the change in energy between successive states, according
to \eqref{piAGibbs}. It is however cleat that $\P(\tau =k)>0$, hence all states are accessible
from a given one, which is equivalent to the irreducibility condition for the Markov chain.

\smallskip

For a single parameter, the set $\Sigma_p$ of possible states has $\# \Sigma_p = 2^{\# V(G)}$. 
The number of states $S_p'$ that are accessible from a given state $S_p$, in a single-spin-flip dynamics, 
is the number of states that differ from $S_p$ by flipping a single spin at one of the vertices. Hence,
for any given $S_p$, there are $N=\# V(G)$ accessible states $S_p'$. The probability of choosing 
one of these states is uniform $\pi(S_p\to S_p')=1/N$. For a given state $S_p$, we have
$\P(S_p\to S_p)=0$ if and only if all the accessible states $S_p'$ with $\pi(S_p\to S_p')=1/N>0$
have lower energy, so that all $\pi_A(S_p\to S_p')=1$ and 
$\P(S_p\to S_p)=1- \sum_{S_p'}\P(S_p\to S_p')=1- \frac{1}{N}\# \{ S_p' \}=0$. Thus, states
$S$ with $\P(S_p\to S_p)=0$ are maxima of the energy. All other states $S_p$
have $\P(S_p\to S_p)>0$, hence they are aperiodic states of the Markov chain. 
If a Markov chain is irreducible then all states have same period, hence the remaining
states must also be aperiodic. Together with irreducibility this implies ergodicity.

\smallskip

Thus, if we focus on the behavior of a single syntactic parameter, we simply have
a Ising model on our graph of languages $G$, with given interaction energies, and
we apply the Metropolis--Hastings algorithm as above.
When we consider more than one parameter at the same time, so that states
are of the form $\vec{S_p}$, under the simplifying assumption that different syntactic
parameters are independent of each other, we just run the same kind of 
Metropolis--Hastings algorithm on each parameter, by each time flipping a single
spin $S_p$ of a single parameter $p$. Namely,
we proceed as described above, using the Metropolis--Hastings algorithm 
with single-spin-flip dynamics.
We propagate a state $s_t\rightarrow s_{t+1}$ by proposing
a configuration $|\vec{S_p}\rangle$ and accepting it, $s_{t+1}=|\vec{S_p}\rangle$, or
rejecting it, $s_{t+1}=s_t$, with some probability. To simplify the noise, we
assume it acts by trying to flip the spin at each vertex independently, although in reality 
this is more likely to occur in clusters. At each step, the algorithm chooses a vertex 
at random and tries to flip its spin, accepting
or rejecting the flip according to \eqref{piAGibbs}:
\begin{equation}\label{accept}
\pi_A(s_t \to s_{t+1}=|\vec{S_p}\rangle)=\left\{
\begin{matrix}
1 & \text{ if }\Delta H_p\le 0 \\
\exp(-\beta\Delta H_p) & \text{ if }\Delta H_p>0.
\end{matrix}\right.
\end{equation}

\smallskip

Propagating the state for many steps produces
a distribution of configurations that asymptotically approaches equation~\eqref{pdf}, 
allowing one to calculate thermal averages of various functions (such as magnetization) 
by evaluating them on the state at each step and averaging over steps.

\subsection{Model with Entailment}

We now discuss how one can adapt the previous computational setting to the
more refined model that also takes into consideration dependencies and
entailment of syntactic parameters.

\smallskip

The first modification in the model consists of the fact that we use, for each
parameter, a set of three possible values $\{ -1,0,+1\}$ instead of the binary
values $\{ \pm 1 \}$, thus allowing for the possibility that a parameter may
be undefined (by effect of the values of other parameters). Thus, instead of
the Ising model, we are looking at a Potts model with $q=3$. The natural
generalization of the Metropolis dynamics of the Ising model is the wider class 
of Glauber dynamics on graphs, where
the single-spin-flip dynamics is replaced by the assumption that each move
changes the parameter value at just one vertex, by choosing a possible value
uniformly at random.  

\smallskip

Consider again the situation described in \S \ref{entsec1}, where we have two parameters
$(p_1,p_2)$ with the property that, if the parameter $p_1=+1$ then the parameter
$p_2$ can take values $\pm 1$, while if the parameter $p_1=-1$ then $p_2$ is undefined.
As before, we associate spin variables $S_{p_1}$ and $S_{p_2}$ to the two parameters,
but now we assume that $S_{p_1}$ has two possible spin states $\{ \pm 1 \}$, as before,
while $S_{p_2}$ has three possible spin states, $\{ +1,0,-1 \}$, with $0$ for the case when 
the parameter is undefined. 

\smallskip

Thus, we see that, in order to accommodate the entailment relations, we need to
construct a multivariable spin glass model, where some of the spin variables 
associated to the vertices behave like an Ising model (with number of spin states
$q=2$), while other variables behave like a Potts model (with number of spin
states $q=3$). 

\smallskip

Moreover, the entailment relation should be modeled by additional interaction terms 
in the Hamiltonian, that couple the variables $S_{p_1}$ and $S_{p_2}$, in such a
way that the configurations with $S_{p_1}=1$ and $S_{p_2}=\pm 1$, and with
$S_{p_1}=-1$ and $S_{p_2}=0$ are favored energetically over all the other possible
combinations. 

\smallskip

This leads us into a new class of spin glass model, which to our knowledge has
not been considered in physics so far, where different spin variables, with different
numbers of spin states coexist on the same graph and are coupled with one another.
We describe a possible choice of an interaction term that achieves the desired result.

\smallskip

We focus on a case with only two parameters $p_1,p_2$ with an entailment
relation as above. Let $S_{\ell,p}$ be the spin variables, as before, where
$S_{\ell,p_1}\in \{ \pm 1 \}$ and $S_{\ell,p_2}\in \{ \pm 1, 0 \}$. We consider
a change of variable for the first spin, with the new variable $X_{\ell,p_1}\in \{ 0,1 \}$
defined by $S_{\ell, p_1}=\exp(\pi i X_{\ell, p_1})$. This just corresponds to the
two possible choices of representing the group $\Z/2\Z$ in the additive form $\{ 0, 1\}$
or in the multiplicative form $\{ +1,-1 \}$. We also associate to the spin variables
$S_{\ell,p_2}\in \{ +1, 0, -1 \}$ a variable $Y_{\ell,p_2}=|S_{\ell, p_2}|\in \{ 0,1 \}$.

\smallskip

We then consider a Hamiltonian given by the sum of two terms
$H = H_E + H_V$, where the term $H_E$ expresses the edge interactions
between different languages, as in \eqref{Hindep}, written in the Potts model form \eqref{HamPottsB}
\begin{equation}\label{HE}
H_E = H_{p_1} + H_{p_2}=- \sum_{\ell,\ell'\in\text{languages}} J_{\ell\ell'}\, \left( \delta_{S_{\ell,p_1} , S_{\ell',p_1}} + \delta_{S_{\ell, p_2} , S_{\ell', p_2}} \right),
\end{equation}
where we treat the two parameters as independent, while the interaction between the parameters
due to the entailment relation is encoded in the $H_V$ term as
\begin{equation}\label{HV}
H_V =  \sum_\ell H_{V,\ell}= \sum_\ell J_\ell \, \delta_{X_{\ell, p_1} , Y_{\ell,p_2}}.
\end{equation}
For $J_\ell>0$, this gives an anti-ferromagnetic pairing between the variables $X_{\ell,p_1}$
and $Y_{\ell,p_2}$. 

\smallskip

The preferred energy states for the Hamiltonian $H_V$
are those where either $X_{\ell,p_1}=0$ and $Y_{\ell,p_2}=1$, or $X_{\ell,p_1}=1$
and $Y_{\ell,p_2}=0$, which correspond, respectively, to the entailment relations
$S_{\ell,p_1}=1$ and $S_{\ell, p_2}=\pm 1$, or $S_{\ell,p_1}=-1$ and $S_{\ell,p_2}=0$. 
Notice that the size of the parameters $J_\ell$ in the term $H_V$ in the Hamiltonian
determine how strongly enforced the entailment relation is in a given language: 
for $J_\ell \to \infty$, an infinite energy barrier separates the ground states $H_{V,\ell}=0$ from
the excited states $H_{V,\ell}= J_\ell$, hence the entailment is strongly enforced, while
lower values of $J_\ell$ would imply that the language $\ell$ admits a certain frequency of
exceptions to this syntactic rule. These should be interpreted in the same sense as the
probabilistic approach to setting syntactic parameters discussed in \cite{Liu}, see also
the discussion in \S \ref{tempSec} above. 

\smallskip

Notice that, if we freeze one of the parameter $p_1$ and only consider the
evolution of the second parameter, then the Hamiltonian above can be regarded
simply as a case of Potts model with external magnetic field. 

\smallskip

We then run the analog of the Metropolis--Hastings algorithm for the Hamiltonian
$H=H_E+H_V$ on the graph of languages and language interactions. Since we
now have spin variables with more than two possible states, instead of the
single-spin-flip dynamics one uses the appropriate Glauber dynamics. For
general results about ergodicity and convergence time of Glauber dynamics
on some classes of graphs, we refer the reader to \cite{Berg}, \cite{Cuff}.
In the simple case we analyze in \S \ref{entailSec} below, we will work with
a complete graph, so the results of \cite{Cuff} apply.

\medskip
\subsection{Implementation}

Part of the implementation consisted of parsing and converting the SSWL database
of syntactic parameters in a form usable for data analysis. Additionally, 
discrepancies in nomenclature between the MIT Media Lab topology source and 
the SSWL database had to be identified and manually resolved. 
The implementation for the independent parameters simulation was executed in MATLAB. 
The implementation for the entailment of parameters was executed in Java.
The source files of the code are available at the GitHub repository: \verb!https://github.com/pointofnoreturn/spin_glass_model! 

\section{Independent Parameters Dynamics}

As a first simulation, we consider all parameters as independent spin variables, hence
we can focus on just one parameter at a time and run a single-parameter simulation.
We report here the case of the Subject-Verb parameter as an illustrative example: we
describe the behavior at low and high temperature $T$, averaging over a million steps.
This is enough to obtain a clear understanding of the phase diagrams of simulations
for all of the other parameters, when considered independently. 
The code in the GitHub repository can be used to generate analogous simulations for
all parameters.

\smallskip

Here, when we refer to the ``low'' and ``high'' temperature regimes, we mean the
range in which the ratio $T/\langle J_{\ell\ell'}\rangle$ is very small or, respectively, very large.

\subsection{Subject-Verb Parameter}

The initial state specified by the
SSWL is given in Figure \ref{init}, where the vertices are represented 
as circular nodes colored red if the language possesses the parameter 
or blue if it lacks the parameter. The sizes of the vertices do not signify anything
in these graphs: variable sizes are only used for ease of visualization. 

\smallskip

Figure \ref{init} suggests that the vast majority of languages currently possess this parameter in the activated
form $+1$. Evolving this system at very low temperature  ($T = 0.000001$), we want to know the
local magnetization, or the thermal average of the spin value at each vertex. Since we are in the low
temperature regime, we expect that the vertices will tend to orient their spins in the same direction
(either all of the languages will tend to possess the parameter or will lack it). Marking vertices
with $\langle S_{\ell,p}\rangle_H>0$ red and vertices with $\langle S_{\ell,p}\rangle_H<0$ blue, we
can look at a graph of these languages at equilibrium, given in Figure \ref{smallt}.

\smallskip

As expected, since the system is close to $T = 0$, it is also close to the ground state. In fact, a configuration
randomly sampled from the equilibrium ensemble will also look like Figure \ref{smallt} nearly 100\% of the time, 
because of how dominant the all-up configuration is: it is nearly impossible at low $T$ for the state to propagate
to a configuration that is not the ground state. At $T = 0$, this is the only configuration in the
ensemble. At this low temperature, the languages quickly converge during their evolution to a state in
which they all acquire the parameter in the activated form $+1$. A more physical way to view this is 
to look at a plot of the average spin as a function of time, as given in the first graph of
Figure \ref{avgspinFig}. 

\smallskip

One may wonder why this system always ends up in this ``all-up'' state instead of an ``all-down'' state.
The key is that the initial configuration is dominated by spin-up vertices, so the energy barrier is much higher
for a droplet of spin-down to expand and dominate the system than it is for the sea of spin-up to swallow
the droplets. 

\smallskip

What happens when $T$ is very large? Taking $T = 20$, the local magnetizations approach zero,
with exactly half of the vertices approaching zero from the positive direction as in Figure \ref{larget}. In fact, the
local magnetizations vary with mean $-2.3927 \times 10^{-4}$ and median $7 \times 10^{-6}$ on the interval $[-0.0186,0.0147]$.

\smallskip

In this case, a configuration picked at random from the equilibrium ensemble is most likely to have 
approximately half of the vertices with spin up. Because $T$ is large, a state can propagate to almost any other
configuration fairly easily. At equilibrium, the languages here evolve approximately independently of each
other, so the average (over vertices) spin fluctuates about zero, as in the second graph of Figure 
\ref{avgspinFig}.

\smallskip

There are then two points to note: the model behaves as predicted, and it is marked 
by the same symmetry-breaking phase transition observed in the 2D Ising Ferromagnet 
at a critical temperature. This is not
surprising, since this model is at its core just a ferromagnet with a more complicated topology.

\section{Model with Entailment of Parameters}\label{entailSec}

We consider here a simple case where we focus on two parameters with an
entailment relation, over a small set of Indo-European languages. In this case, 
we take the data on the values of the syntactic parameters from \cite{Longo1}, \cite{Longo2},
while we still use the same data from the MIT Media Lab database for the
interaction strengths between different languages. In this simulation we used the
Media Lab data based on book translations instead of Wikipedia editing\footnote{{\tt http://language.media.mit.edu/visualizations/books}}.

\begin{figure}
\begin{center}
\includegraphics[scale=0.65]{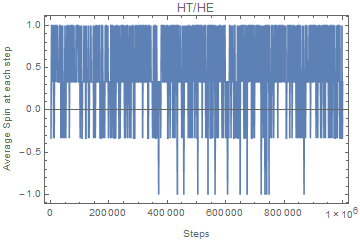}
\includegraphics[scale=0.65]{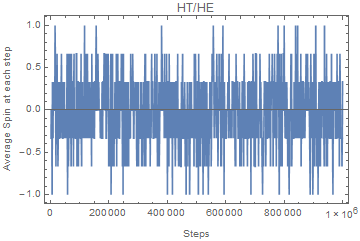}
\includegraphics[scale=0.65]{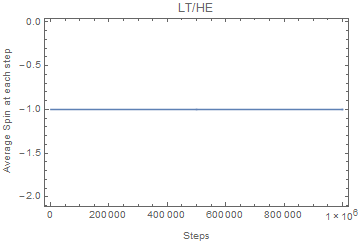}
\includegraphics[scale=0.65]{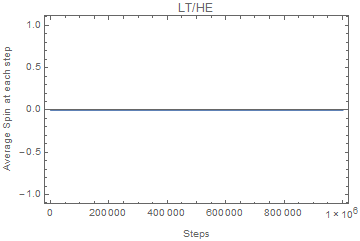}
\end{center}
\caption{Average value of spin for $p_1=\text{Partial Definiteness}$ 
(left) and for $p_2=\text{Definiteness Checking}$ (right) in
the high temperature/high energy (HT/HE) regime (top) and
in the low temperature/high energy (LT/HE) regime (bottom), 
as a function of the number of steps in the Monte Carlo simulation.}
\label{avgspinp1p2G1HTHELTHEFig}
\end{figure}

\begin{figure}
\begin{center}
\includegraphics[scale=0.65]{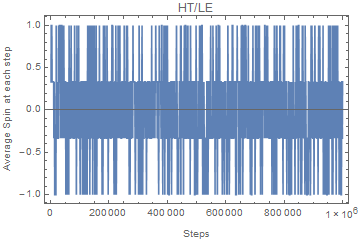}
\includegraphics[scale=0.65]{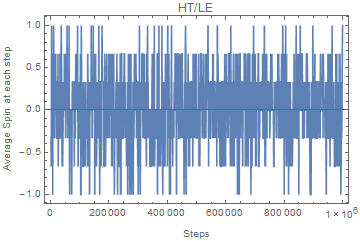}
\includegraphics[scale=0.65]{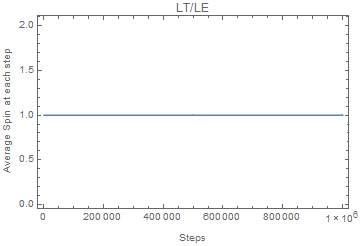}
\includegraphics[scale=0.65]{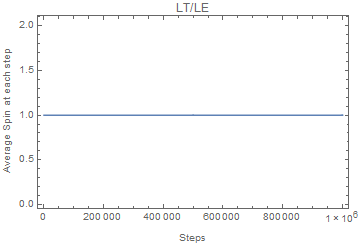}
\end{center}
\caption{Average value of spin for $p_1=\text{Partial Definiteness}$ 
(left) and for $p_2=\text{Definiteness Checking}$ (right) in
the high temperature/low energy (HT/LE) regime (top) and
in the low temperature/low energy (LT/LE) regime (bottom), 
as a function of the number of steps.}
\label{avgspinp1p2G1HTLELTLEFig}
\end{figure}

\begin{figure}
\begin{center}
\includegraphics[scale=0.65]{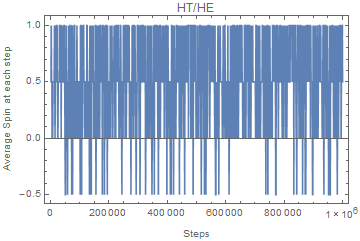}
\includegraphics[scale=0.65]{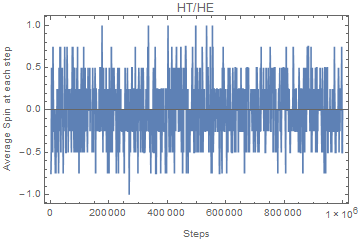}
\includegraphics[scale=0.65]{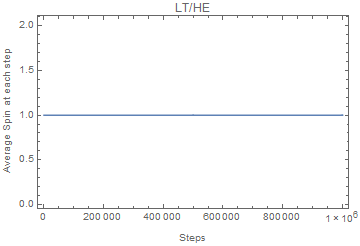}
\includegraphics[scale=0.65]{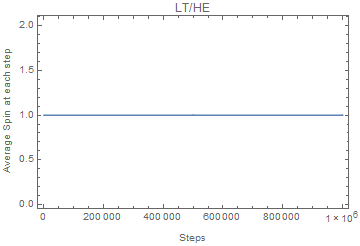}
\end{center}
\caption{Average value of spin for $p_1=\text{Strong Deixis}$  
(left) and for $p_2=\text{Strong Anaphoricity}$  (right) in
the HT/HE regime (top) and the LT/HE regime (bottom), 
as a function of the number of steps.}
\label{avgspinp1p2G2HTHELTHEFig}
\end{figure}

\begin{figure}
\begin{center}
\includegraphics[scale=0.65]{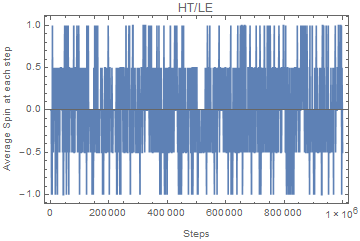}
\includegraphics[scale=0.65]{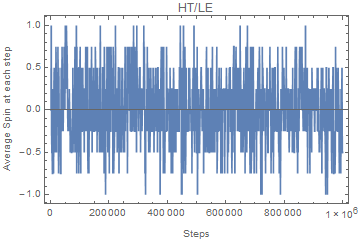}
\includegraphics[scale=0.65]{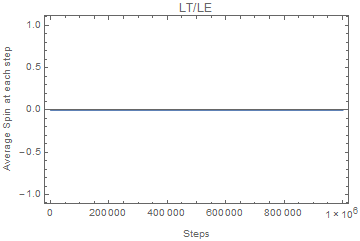}
\includegraphics[scale=0.65]{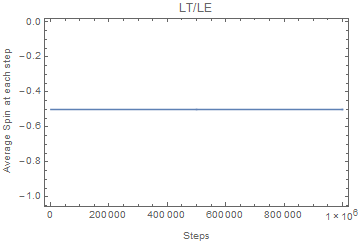}
\end{center}
\caption{Average value of spin for $p_1=\text{Strong Deixis}$  
(left) and for $p_2=\text{Strong Anaphoricity}$  (right) in
the HT/LE regime (top) and the LT/LE regime (bottom), 
as a function of the number of steps.}
\label{avgspinp1p2G2HTLELTLEFig}
\end{figure}

\smallskip

We consider, for example two parameters $p_1,p_2$ expressing Definiteness 
(respectively, number~7 and 12 in the list in Table A of \cite{Longo1}). The
first parameter $p_1$ expresses Partial Definiteness, while the parameter $p_2$
is Definiteness Checking. See for instance \cite{Corn} for a discussion of the
role of these parameters. We consider three languages: English, Russian, and
Bulgarian.  We input interaction energies taken from the same MIT Media Lab data graph,
and we neglect the effect of interaction with other languages. 
The initial configuration of the two parameters, taken 
from Table A of \cite{Longo1} is according to the following table:
\begin{center}
\begin{tabular}{|c|c|c|}
\hline
              & $p_1$ & $p_2$ \\
\hline
$\ell_1$ & $+1$ & $-1$  \\
\hline
$\ell_2$ & $-1$ & $0$  \\
\hline
$\ell_3$ & $+1$ & $+1$  \\
\hline
\end{tabular}
\end{center}

\smallskip

Our second example considers another pair of parameters $p_1,p_2$,
given, respectively, by Strong Deixis and Strong Anaphoricity (numbers 52 and 53 in
Table A of \cite{Longo1}). For a detailed discussion of deixis see \cite{Lenz}, while
regarding anaphoricity, see for instance \cite{Cole}.
In this example, we work with the complete graph
on four vertices (tetrahedron graph), where the vertices $\ell_1,\ldots, \ell_4$
correspond to the languages: English, Welsh, Russian, and Bulgarian.
Notice that, according to the Media Lab graph, the interaction energies between the pairs 
Welsh/Russian and Welsh/Bulgarian are negligible compared to the other
interaction energies. 
The initial configuration, from Table A of \cite{Longo1}, for this
pair of parameters and these four languages is the following:
\begin{center}
\begin{tabular}{|c|c|c|}
\hline
              & $p_1$ & $p_2$ \\
\hline
$\ell_1$ & $+1$ & $+1$  \\
\hline
$\ell_2$ & $-1$ & $0$  \\
\hline
$\ell_3$ & $+1$ & $+1$  \\
\hline
$\ell_4$ & $+1$ & $-1$ \\
\hline
\end{tabular}
\end{center}


\medskip
\subsection{Entailment dynamics}

We ran a simulation of the dynamics of the system for two entailed parameters 
$p_1,p_2$, with Hamiltonian $H=H_E +H_V$, where $H_E$ and $H_V$ are
given, respectively, by \eqref{HE} and \eqref{HV}, in the case of the two
examples described above. 

\smallskip

In this simulation, we replaced the acceptance probabilities \eqref{piAGibbs} with
\begin{equation}\label{piAGibbs2}
\pi_A(s\rightarrow s\pm1\, ({\rm mod}\,3))=\left\{\begin{array}{ll}
1&\text{ if } \Delta_H \le 0 \\
\exp(-\beta \Delta_H) & \text{ if }\Delta_H >0.
\end{array}\right.
\end{equation}
where
$$ \Delta_H := \min\{H(s+1\, ({\rm mod}\,3)),H(s-1 \, ({\rm mod}\,3))\}-H(s). $$

\smallskip

The dynamics depends on two parameters, the temperature and the coupling
energy of the entailment relation. 
We ran simulations for high/low temperature
and high/low entailment energy. 

\smallskip

In the first case, for the graph with languages $\{ \ell_1, \ell_2, \ell_3 \}=\{ \text{English}, \text{Russian},
\text{Bulgarian} \}$ and the parameters $\{ p_1, p_2 \}=\{ \text{Partial Definiteness}, 
\text{Definiteness Checking}\}$, we consider an initial state as given in the first table above.
The average value of spin for the two parameters is illustrated in Figures~\ref{avgspinp1p2G1HTHELTHEFig}
and \ref{avgspinp1p2G1HTLELTLEFig},  in the different regimes of high temperature and high entailment
energy (HT/HE), high temperature and low entailment energy (HT/LE), low temperature and high entailment
energy (LT/HE), low temperature and low entailment energy (LT/LE). The final equilibrium 
states for the dynamics, in each of these cases, would then be as follows:
\bigskip
\begin{center}
\begin{tabular}{|c||c|c|c|c|}
\hline
 $(p_1,p_2)$  & HT/HE & HT/LE & LT/HE & LT/LE \\
\hline \hline
$\ell_1$          & $( +1,+1 )$  & $(+1 ,+1 )$  &  $( -1,0 )$ &  $( +1,+1 )$ \\
\hline
$\ell_2$          & $(+1 ,0 )$   &  $(+1 , +1 )$  &  $(-1 ,0 )$ &  $( +1,+1 )$ \\
\hline
$\ell_3$          & $(-1 , 0 )$   &  $( -1,+1 )$  &  $(-1 ,0 )$ &  $(+1 ,+1 )$ \\
\hline
\end{tabular}
\end{center}

\bigskip
\newpage

In the second case, with languages $\{ \ell_1,\ell_2, \ell_3,\ell_4 \}=\{ 
\text{English}, \text{Welsh}, \text{Russian}, \text{Bulgarian} \}$ and with parameters
$\{ p_1, p_2 \} = \{ \text{Strong Deixis}, \text{Strong Anaphoricity} \}$, the initial state
is given by the values of the parameters in the second table above, and the 
average value of spin in the different HT/HE, HT/LE, LT/HE, LT/LE regimes is
illustrated in Figures~\ref{avgspinp1p2G2HTHELTHEFig} and
\ref{avgspinp1p2G2HTLELTLEFig}.
The final equilibrium states for the dynamics are then
of the form
\bigskip
\begin{center}
\begin{tabular}{|c||c|c|c|c|}
\hline
 $(p_1,p_2)$  & HT/HE & HT/LE & LT/HE & LT/LE \\
\hline \hline
$\ell_1$          & $( +1 ,0  )$  & $( +1,-1 )$  &  $(  +1,+1  )$ &  $( +1 , -1 )$ \\
\hline
$\ell_2$          & $( +1 ,-1  )$   &  $( -1 ,-1)$  &  $( +1 ,+1  )$ &  $( +1 , -1  )$ \\
\hline
$\ell_3$          & $( -1 , 0  )$   &  $( -1 , +1)$  &  $( +1 , +1 )$ &  $(-1 , 0 )$ \\
\hline
$\ell_4$          & $(  +1,+1)$   &  $( -1, -1)$  &  $( +1 ,+1  )$ &  $( -1 , 0 )$ \\
\hline
\end{tabular}
\end{center}

\bigskip

These examples are only illustrative and not entirely realistic, because we have
singled out a small portion of the language graph that exhibits an interesting
configuration of entailed parameters in the initial state, and we have run the
simulation neglecting the interactions with all the other languages in the rest
of the larger language graphs, which will also affect the behavior of the system.
These examples, however, are interesting because they show situations where
a configuration of entailed parameters reaches an equilibrium states where
parameters of the individual languages have undergone some changes,
but have not always converged to a configuration where all the parameters
are aligned. While in the first example one obtains complete alignment of
all the parameters in the low temperature and low energy regime, in the
second example, even in this range, parameters do not fully align. This shows
that the presence of entailment between syntactic parameters can have a
substantial effect on the dynamics that differs significantly in outcome from
the case where one assumes an independence hypothesis on parameters.

\section{Conclusions and further questions}

In this paper, we introduced a new tool to study the flow of syntactic parameters by mapping the
problem onto a spin glass model. This tool provides some flexibility, as one could provide any 
topology and any adjacency matrix. 
We have shown that, under a hypothesis of independence between syntactic parameters,
in the evolution of the system parameters tend to align to the $+1$ position. We related this
to linguistic models of bilingual code-switching. We also showed that, when entailment
relations between parameters are taken into account, the uncoupled Ising models are
replaced by a coupling at the vertices of Ising and Potts models with same edge interactions.
One obtains then more complicated equilibrium configurations, depending on the entailment 
energy parameter, which gives the strength of the coupling, and its
relation to the ``temperature" parameter. Even in the low energy and low temperature
regime one finds cases that do not yield a complete alignment of parameters. 

\smallskip

One factor not accounted for in this project was the
number of speakers for each language, which could significantly change the interaction strengths. Another
interesting scenario would involve disordering terms in which two interacting languages actually prefer to
antialign their parameters.

\smallskip

An overall hypothesis of this model is that the interaction energies along the edges, 
which are modeled on estimates of the size of the bilingual population, remain constant during
the evolution. In a realistic model, the strength of the interaction would
also be co-evolving. It is also likely that factors such as the existence of better and more
efficient automated translators will eventually cause drastic changes in the size of human
bilingual or multilingual populations, so the feasibility of such models of linguistic
evolution may also be altered by external factors of this sort.


\subsection*{Acknowledgment} This work was performed as part of the activities of the last author's 
Mathematical and Computational Linguistics lab and CS101/Ma191 class at Caltech. The last author 
is partially supported by NSF grants DMS-1201512 and PHY-1205440.

\end{document}